\documentclass[10pt,twocolumn,letterpaper]{article}

\usepackage{cvpr}              
\definecolor{cvprblue}{rgb}{0.21,0.49,0.74}
\usepackage[pagebackref,breaklinks,colorlinks,allcolors=cvprblue]{hyperref}
\usepackage[accsupp]{axessibility}  

\usepackage{algorithmic}
\usepackage{graphicx}
\usepackage{tabularx}
\usepackage{makecell}
\usepackage{xcolor}
\usepackage{bm}
\usepackage{amsmath} 
\usepackage{multirow}
\usepackage{algorithm}

\usepackage{amssymb}

\def\paperID{30896} 
\def\confName{CVPR}
\def\confYear{2026}

\title{CCCaption: Dual-Reward Reinforcement Learning for Complete and Correct Image Captioning}
\author{Zhijiang Tang$^{1,2*}$, Linhua Wang$^{3*}$, Jiaxin Qi$^{1,2*}$, Weihao Jiang$^{3}$, 
\\
Peng Hou$^{3}$, Anxiang Zeng$^{3\dagger }$, Jianqiang Huang$^{1,2\dagger}$
\\
$^1$Computer Network Information Center, Chinese Academy of Sciences, Beijing, China\\
$^2$Hangzhou Institute for Advanced Study, University of Chinese Academy of Sciences, Zhejiang, China\\
$^3$LLM Team, Shopee Pte. Ltd., Shanghai, China\\
{\tt\small tangzhijiang24@mails.ucas.ac.cn, \{jxqi, jqhuang\}@cnic.cn,}\\
{\tt\small \{wanglinhua, weihao.jiang, peng.hou, anxiang.zeng\}@shopee.com}
}

\begin{document}
\maketitle
\footnotetext[1]{Equal contribution. }
\footnotetext[2]{Corresponding authors.}
\begin{abstract}

Image captioning remains a fundamental task for vision–language understanding, yet ground-truth supervision still relies predominantly on human-annotated references.
Because human annotations reflect subjective preferences and expertise, ground-truth captions are often incomplete or even incorrect, which in turn limits caption models.
We argue that caption quality should be assessed by two objective aspects: completeness (does the caption cover all salient visual facts?) and correctness (are the descriptions true with respect to the image?).
To this end, we introduce CCCaption: a dual-reward reinforcement learning framework with a dedicated fine-tuning corpus that explicitly optimizes these properties to generate \textbf{C}omplete and \textbf{C}orrect \textbf{Captions}.
For completeness, we use diverse LVLMs to disentangle the image into a set of visual queries, and reward captions that answer more of these queries, with a dynamic query sampling strategy to improve training efficiency.
For correctness, we penalize captions that contain hallucinations by validating the authenticity of sub-caption queries, which are derived from the caption decomposition.
Our symmetric dual-reward optimization jointly maximizes completeness and correctness, guiding models toward captions that better satisfy these objective criteria. Extensive experiments across standard captioning benchmarks show consistent improvements, offering a principled path to training caption models beyond human-annotation imitation.

\end{abstract} 
\vspace{-12pt}
\section{Introduction}
\label{sec:intro}

Image captioning requires models to generate faithful, comprehensive linguistic descriptions of images, and serves as a core interface between vision and language, bridging the modality gap~\cite{showandtell, lu2019vilbert, clip, qwen3}.
Captioning also matters for pretraining multimodal large language models (MLLMs), which are trained on web image–caption pairs~\cite{yin2024survey, llava}, and are often sparse or even biased. Thus, captioning models can re-caption these data and enhance their quality, thereby improving MLLMs' pretraining~\cite{nguyen2023improving, lai2024revisit}.

\begin{figure}
    \centering
    \includegraphics[width=\linewidth]{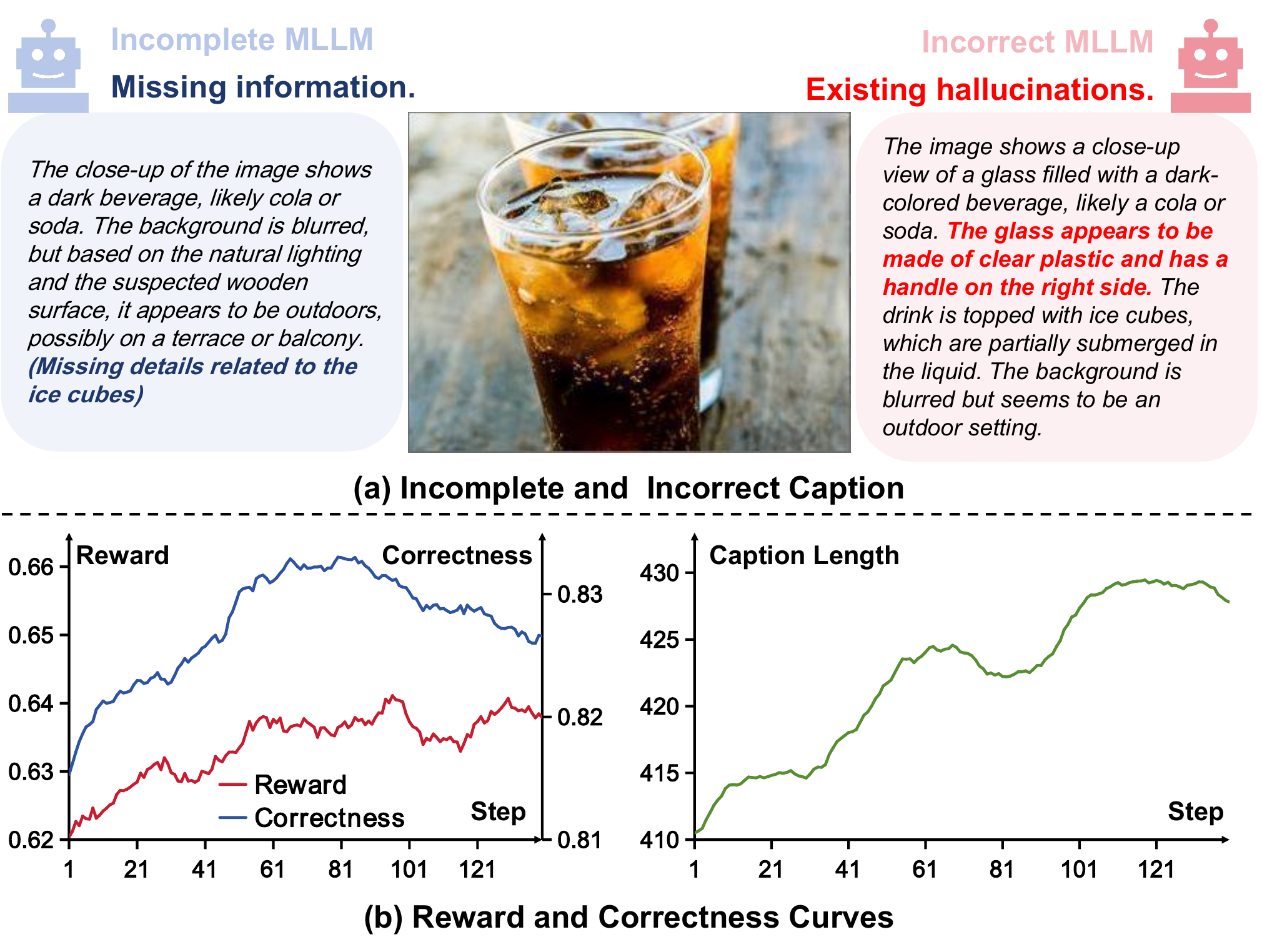}
    \caption{(a) Comparison of incomplete and incorrect captions generated by MLLMs. An example of incomplete caption with missing information versus an incorrect caption with hallucinated details. (b) Reward and correctness curves during model training, illustrating the progression of reward, correctness (as defined in Section~\ref{method}), and caption length across training steps.
}
    \label{fig:teaser}
    \vspace{-12pt}
\end{figure}

However, previous captioning models heavily rely on human-annotated references for training~\cite{feng2019unsupervised, blip}, and these annotations often reflect annotators’ preferences and expertise, and models that merely mimic these annotations cannot reliably produce complete, objective captions~\cite{wang2021human}. 
Analyzing the nature of captions, we argue that a high-quality caption must satisfy two fundamental properties: completeness and correctness. As illustrated in Figure~\ref{fig:teaser}(a), completeness requires covering the salient visual facts (e.g., the left caption is incomplete because it omits the ice cubes). Correctness demands image-grounded factuality (e.g., the right caption mentions key content but misidentifies the container as a plastic cup with a handle, thus violating correctness).

To optimize these objective properties, recent work adopts reinforcement learning to optimize task-aligned objectives~\cite{guo2025deepseek, ren2017deep}. For example, CapRL~\cite{caprl} rewards captions that answer more visual queries to achieve completeness. 
However, it derives training queries from a single MLLM, limiting coverage of visual content and yielding incomplete supervision. Moreover, the unconstrained reward may encourage the model to describe more image details to maximize the reward, which often leads to more hallucinations~\cite{lee2024toward, feng2024more}.
As shown in Figure~\ref{fig:teaser}(b), caption correctness drops markedly in later training stages even as the reward keeps increasing. This divergence suggests that captions learn to answer queries while hallucinating off-query details that the queries do not constrain, thereby inflating reward yet degrading factuality. Thus, a robust captioning objective should enforce completeness and correctness jointly.



To address the limitations of existing methods~\cite{phimsiri2025trafficinternvl, caprl} and achieve more complete and correct captions, we propose \textbf{CCCaption}, a reinforcement learning framework that explicitly combines \textbf{C}ompleteness and \textbf{C}orrectness rewards to optimize \textbf{Caption}ing models.

(1) For \textit{completeness}, we reward captions that answer more queries. A multi-MLLMs query generation pipeline was developed to enhance the query coverage and make it more comprehensive. Firstly, we use diverse MLLMs to decompose the image into a series of visual queries. By introducing a diversity measure~(i.e., the greater the variance of queries, the greater the diversity of queries) for the query set with additional rules (e.g., query validity and relevance), we filter out low-quality queries. Through repeated iterations, complete queries that fully capture the image's information are obtained. Finally, a \textbf{C}omplete \textbf{Caption} training dataset consisting of 44k samples, \textbf{CCaption-44k} was constructed. In addition, we designed a dynamic query sampling strategy during training, which increases the sampling probability of diverse queries (i.e., queries with higher advantages) to improve training efficiency.

(2) For \textit{correctness}, we penalize captions that contain hallucinations. Building on previous work~\cite{ye2025painting, lee2024toward}, we decompose a long caption into sub-caption queries, which include the basic information of the caption.
Then, input the atomic queries along with the original image into a MLLM, allowing the model to score hallucinations in the atomic queries and thereby providing a correctness reward for the caption. So our CCCaption reinforcement learning framework hybridizes the completeness and correctness rewards, along with CCaption-44k training data, and addresses both completeness and correctness of the caption. This allows the captioning model to generate longer and more detailed captions while significantly reducing hallucinations. 

In experiments, we trained a Qwen3-VL-2B~\cite{qwen3} model, \textbf{CCCaption-2B}, using our CCCaption framework. And extensive experiments show that our performance not only surpasses that of Qwen3-VL-32B but also achieves state-of-the-art results in image captioning.

%

We summarize our contributions as follows:
\begin{itemize}
\item We analyse the limitations of existing captioning models and propose a set of criteria for evaluating caption quality: completeness and correctness.
\item Based on these two criteria, we design the CCCaption framework, which combines a dual-reward mechanism for reinforcement learning to generate high-quality captions, along with our comprehensive CCaption-44k. Then we trained a powerful captioning model, CCCaption-2B.
\item We conduct extensive performance experiments, which demonstrate the effectiveness of our proposed framework, particularly in controlling model hallucinations.
\end{itemize}

\noindent\textbf{Code: }\href{https://github.com/Shopee-MUG/CCCaption}{https://github.com/Shopee-MUG/CCCaption}

\section{Related Works}

\subsection{Image Captioning}
Early image captioning methods primarily used the encoder-decoder framework: for instance, Show and Tell with its variant~\cite{vinyals2015show,xu2015show} used convolutional neural networks~\cite{krizhevsky2012imagenet} to extract image features, and then utilized recurrent neural networks~\cite{bengio1994learning} to generate textual descriptions. Image Captioning~\cite{herdade2019image} incorporated object relationship encoding and Transformer~\cite{vaswani2017attention} to enhance the model’s understanding of spatial relationships. More recent methods have also integrated multimodal features and visual relationship modeling~\cite{li2019visualbert, chen2020uniter}. With the rise of multimodal large language models~\cite{li2022blip, liu2023visual, chen2024internvl}, models can now achieve powerful captioning capabilities through supervised fine-tuning~\cite{li2023blip, bucciarelli2024personalizing}. Although these methods have made significant progress on public benchmarks (e.g., MS COCO~\cite{lin2014microsoft}), traditional supervised training still relies on human-annotated reference captions, which reflect the annotators’ preferences and often result in incomplete captions, as well as hallucinations.

\subsection{Reinforcement Learning for Captioning}
To move beyond simply mimicking human annotations, researchers have started integrating reinforcement learning~(RL)~\cite{watkins1992q, shao2024deepseekmath} into captioning models. Early work~\cite{ren2017deep} employed an actor-critic architecture and proposed a visual-semantic embedding reward. Later, methods such as~\cite{liu2018multi} introduced multi-level policies and multimodal reward functions to optimize the caption generation in greater detail. The latest research, such as CapRL~\cite{caprl}, further integrates “question-answering ability” as a metric for caption quality to avoid reward hacking introduced by MLLMs as judges. Despite RL’s potential to improve caption quality, it has introduced new challenges, such as low training efficiency, overly verbose descriptions, and severe hallucinations. In response, we propose the CCCaption framework, a hybrid reward mechanism that combines completeness and correctness with an efficient query sampling strategy to achieve breakthroughs in RL-based captioning.

\begin{figure*}
    \centering
    \includegraphics[width=\linewidth]{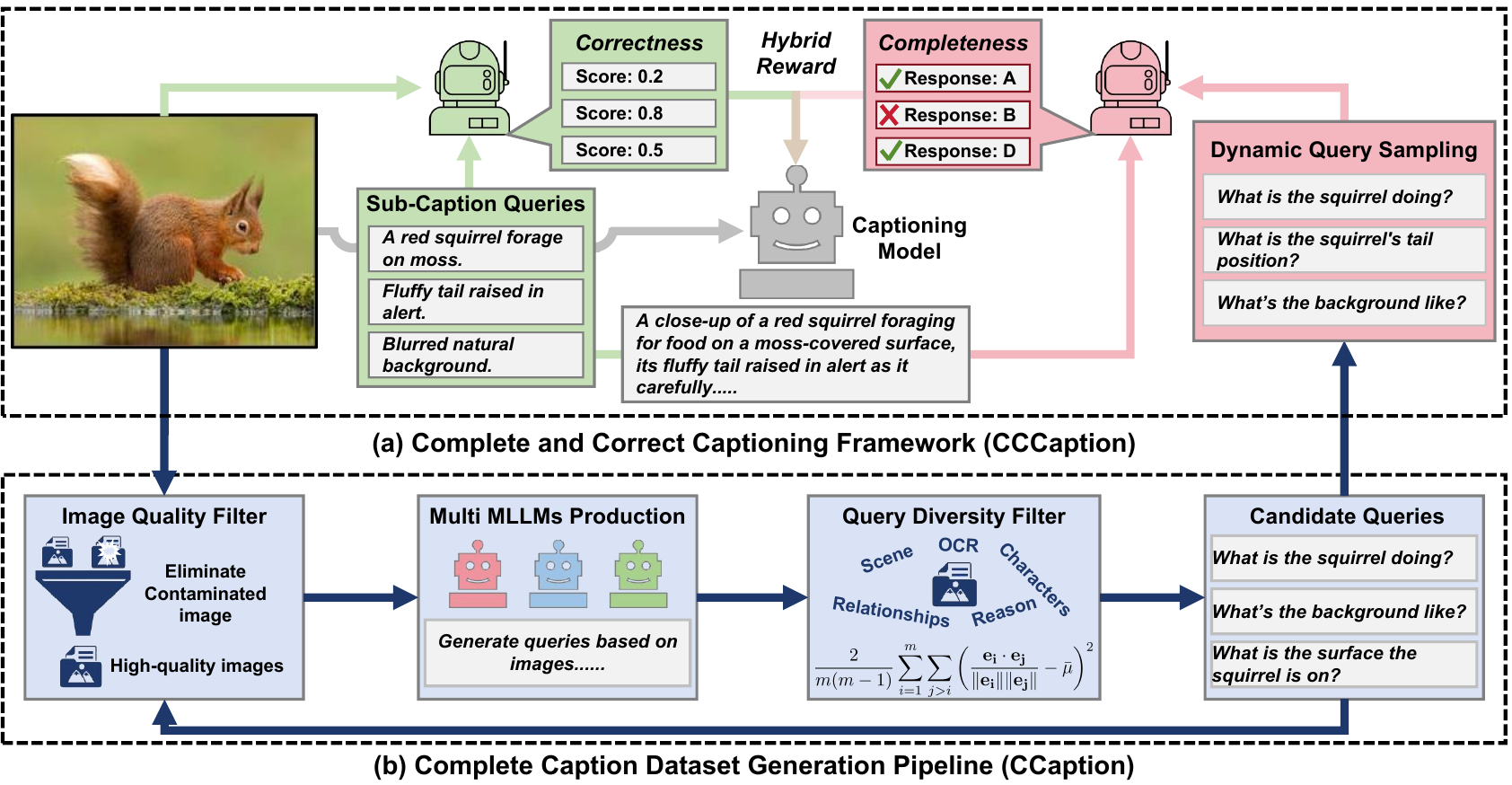}
    \caption{Illustrations of our Complete and Correct Captioning (CCCaption) framework and Complete Caption dataset (CCaption) generation pipeline. (a) The CCCaption reinforcement learning framework. The green and red parts represent the processes for computing the correctness and completeness rewards, respectively. We combine these two rewards using the GRPO algorithm~\cite{guo2025deepseek} to perform reinforcement learning training for the captioning model, while employing the dynamic query sampling strategy to enhance training efficiency. (b) The CCaption dataset generation pipeline. By iterating through the entire process, a diverse, complete query set is generated, ensuring thorough coverage of all the image information.}
    \label{fig:method}
    \vspace{-12pt}
\end{figure*}

\section{Method}
\label{method}
%
\subsection{Preliminaries}


\noindent\textbf{Image Captioning}. Given an image captioning dataset $\mathcal{D} \!=\! \{(\mathbf{x}, \mathbf{y})\}$, 
where $\mathbf{x}$ denotes images and $\mathbf{y}$ denotes captions, 
a captioning model $M_\theta$ defines a conditional distribution 
$p_{\theta}(\mathbf{y}\!\mid\!\mathbf{x})$ and is typically optimized by maximum likelihood estimation with the token-level cross-entropy loss:
\begin{equation}
\label{MLE}
    \mathcal{L}_{\text{MLE}}(\theta)
    = -\frac{1}{|\mathcal{D}|}
    \sum_{(\mathbf{x},\mathbf{y})\in\mathcal{D}}
    \sum_{i=1}^{|\mathbf{y}|}
    \log p_{\theta}\big(y_{i}\mid y_{<i},\mathbf{x}\big),
\end{equation}
where $\mathbf{y} = (y_1, y_2, \ldots)$ is the sequence of reference tokens and $y_{<i}$ denotes the prefix $(y_1,\ldots,y_{i-1})$.


However, when the references are biased or incomplete, a model trained on these annotations cannot generate desirable captions. To move beyond reference imitation, recent work leverages reinforcement learning (RL) to optimize task-specific objectives~\cite{guo2025deepseek, ren2017deep, caprl}. Let $\pi_{\mathcal{D}}$ denote the empirical distribution over images in $\mathcal{D}$ and $\pi_{\theta}(\cdot\mid\mathbf{x})$ denote the caption distribution induced by $M_\theta$. The RL objective can be written as maximizing the expected reward:
\begin{equation}
\label{RL}
    \mathcal{J}_{\text{RL}}(\theta)
    = \mathbb{E}_{\mathbf{x}\sim\pi_{\mathcal{D}}}
      \;\mathbb{E}_{\hat{\mathbf{y}}\sim\pi_{\theta}(\cdot\mid\mathbf{x})}
      \big[\,\mathcal{R}(\mathbf{x},\hat{\mathbf{y}})\,\big],
\end{equation}
where $\hat{\mathbf{y}}$ is a sampled caption for $\mathbf{x}$ and $\mathcal{R}(\mathbf{x},\hat{\mathbf{y}})$ is a scalar reward that evaluates its quality. 
We then instantiate $\mathcal{R}$ to explicitly capture two key aspects of caption quality, \emph{completeness} and \emph{correctness}, and optimize $M_\theta$ accordingly.

\noindent\textbf{Completeness and Correctness}.
Given an image--caption pair $(\mathbf{x}, \mathbf{y})$, we represent their basic information by two sets,
$\mathcal{B}_{\mathbf{x}} \!=\! \{\mathbf{b}_i\}_{i=1}^{k_{\mathbf{x}}}$ for the image and $\mathcal{B}_{\mathbf{y}} \!=\! \{\mathbf{b}_j\}_{j=1}^{k_{\mathbf{y}}}$ for the caption. Each element $\mathbf{b}$ denotes an atomic semantic fact (e.g., objects). Then, the completeness and correctness of caption $\mathbf{y}$ for image $\mathbf{x}$ are quantified by the following scores:
\begin{equation}
    s_{\text{comp}}(\mathbf{x},\mathbf{y})
    = \frac{\bigl|\mathcal{B}_{\mathbf{x}} \cap \mathcal{B}_{\mathbf{y}}|}
           {\bigl|\mathcal{B}_{\mathbf{x}}\bigr|};\;
    s_{\text{corr}}(\mathbf{x},\mathbf{y})
    = \frac{\bigl|\mathcal{B}_{\mathbf{x}} \cap \mathcal{B}_{\mathbf{y}}|}
           {\bigl|\mathcal{B}_{\mathbf{y}}\bigr|}.
\end{equation}
Intuitively, $s_{\text{comp}}$ measures how many image facts are
covered by the caption, while $s_{\text{corr}}$
measures how many caption facts are grounded in the image.
In practice, however, the basic information sets are unobservable and cannot be computed directly. Therefore, we propose tractable approximations in the following sections.

\begin{algorithm}[tb]
\caption{Queries Sampling with Multi-MLLMs}
\label{alg:qa_pipline}
\textbf{Input}: MLLMs \( {M}_{\text{list}} \), queries required \( n_{\text{q}} \), threshold \( \tau \) \\
\textbf{Output}: Query list \( \mathcal{Q}_{\text{list}} \)
\begin{algorithmic}[1]
\FOR{ \( i = 1 \) to Attempts }
    \STATE Randomly select \( {M} \) from \( {M}_{\text{list}} \)
    \STATE \( \bm{q} \gets {M}(\bm{x}, \text{prompt}) \)
    \IF{\( \bm{q}\) is valid and related to image $\bm{x}$}
        \STATE Append \( \bm{q} \) to \( \mathcal{Q}_{\text{list}} \)
    \ELSE
        \STATE Continue
    \ENDIF
    \IF{length of \( \mathcal{Q}_{\text{list}}  < n_{\text{q}} \)}
        \STATE Continue
    \ELSIF{  Diversity \( \mathcal{V}(\mathcal{Q}_{\text{list}}) < \tau \)}
        \STATE Remove least contribution query in $\mathcal{V}$ from \( \mathcal{Q}_{\text{list}} \)
    \ELSE
        \STATE Break
    \ENDIF
\ENDFOR
\STATE \textbf{return} \( \mathcal{Q}_{\text{list}} \)
\end{algorithmic}
\end{algorithm}

\subsection{Our Framework: CCCaption}
To explicitly capture the two key aspects of caption, completeness and correctness, we propose the corresponding rewards to optimize $M_\theta$. In addition, we introduce a dynamic query sampling strategy to improve training efficiency.

\noindent\textbf{Completeness Reward}.
We approximate the basic information set $\mathcal{B}_{\mathbf{x}}$ by a set of visual queries
$\mathcal{Q}_{\mathbf{x}} \!=\! \{\mathbf{q}_i\}_{i=1}^{m}$, and reward captions that can answer more of these queries. Formally, the completeness reward is defined as
\begin{equation}
    \mathcal{R}_{\text{comp}}(\mathbf{x},\hat{\mathbf{y}})
    = \frac{1}{|\mathcal{Q}_{\mathbf{x}}|}
      \sum_{\mathbf{q} \in \mathcal{Q}_{\mathbf{x}}}
      \mathbb{I}\bigl( M_{J}(\hat{\mathbf{y}}, \mathbf{q}) \bigr),
\end{equation}
where $\hat{\mathbf{y}} \!=\! M_\theta(\mathbf{x})$ is the generated caption, $M_{J}$ is a frozen third-party judge model that answers $\mathbf{q}$ based on $\hat{\mathbf{y}}$, and $\mathbb{I}(\cdot)$ is an indicator that returns $1$ when the answer is correct.

Obviously, the more diverse the query set $\mathcal{Q}_{\mathbf{x}}$, the more reliable the completeness reward becomes. Current methods, such as CapRL~\cite{caprl}, generate queries using a single MLLM with simple prompts, which suffer from single-model bias and limited coverage. To alleviate this issue, we define a diversity $\mathcal{V}$ over a query set $\mathcal{Q}_{\mathbf{x}}$:
\begin{equation}
    \mathcal{V}(\mathcal{Q}_\mathbf{x})=\frac{2}{m(m-1)}\sum_{i=1}^m\sum_{j>i}\left (  \frac{\mathbf{e_i}\cdot\mathbf{e_j}}{\|\mathbf{e_i}\|\|\mathbf{e_j}\|}-\bar{\mathbf{\mu}}\right )^2
\end{equation}
where $\mathbf{e}_i$ is the embedding vector of query $\mathbf{q}_i\!\in\!\mathcal{Q}_\mathbf{x}$, 
$m$ represents the size of the query set, i.e., $m = |\mathcal{Q_{\mathbf{x}}}|$.
And $\small{ \mathbf{\bar{\mu}} \!=\! \frac{2}{m(m-1)} \sum_{i=1}^m \sum_{j>i} \frac{\mathbf{e_i} \cdot \mathbf{e_j}}{\|\mathbf{e_i}\| \|\mathbf{e_j}\|} }$. Based on the diversity definition, we design a multi-MLLM query generation algorithm (as shown in Algorithm~\ref{alg:qa_pipline}) that generates and filters queries using multiple MLLMs to promote query diversity and ensure completeness. Applying our algorithm to large-scale images yields a training set of 44k samples, with an average of 10 queries per image, denoted CCaption-44k.

\noindent\textbf{Correctness Reward}. Building on prior work~\cite{ye2025painting, lee2024toward} on evaluating correctness,
we decompose the generated caption $\hat{\mathbf{y}}$ into a set of sub-caption queries $\mathcal{Q}_{\hat{\mathbf{y}}} \!=\! \{\mathbf{q}_i\}_{i=1}^{n}$, which serves as an approximation to the basic information set $\mathcal{B}_{\mathbf{y}}$.
We then reward captions whose sub-caption queries are better grounded in the image.
Formally, the correctness reward is defined as
\begin{equation}
    \mathcal{R}_{\text{corr}}(\mathbf{x},\hat{\mathbf{y}})
    = \frac{1}{|\mathcal{Q}_{\hat{\mathbf{y}}}|}
      \sum_{\mathbf{q} \in \mathcal{Q}_{\hat{\mathbf{y}}}}
      M_{J}(\mathbf{x}, \mathbf{q}),
\end{equation}
where $M_{J}$ is the same frozen third-party judge model, $M_J(\mathbf{x}, \mathbf{q}) \in [0,1]$ provides a soft assessment of how well each caption query is grounded given the image $\mathbf{x}$.


\begin{table}[t!]
  \centering
    \renewcommand{\arraystretch}{1.2}
    \scalebox{0.9}{
\begin{tabularx}{0.53\textwidth}{l|>{\centering\arraybackslash}X>{\centering\arraybackslash}X>{\centering\arraybackslash}X>{\centering\arraybackslash}X>{\centering\arraybackslash}X>{\centering\arraybackslash}X}
    \toprule
    \multicolumn{1}{c|}{\multirow{2}{*}{{MLLM}}}
      & Claude & CPMV & Cog2 & \multicolumn{1}{c}{\multirow{2}{*}{{Avg.}}} \\
      & \cite{anthropic2024claude} &\cite{huggingface2024minicpm} & \cite{huggingface2024cogvlm2} &  \\
    \midrule
    LLaVa-V1.6-34B~\cite{llava_v16} & -35.17  & -4.17  & -26.92  & -22.08  \\
    Qwen2.5-VL-3B~\cite{qwen2_5_vl} & -25.33  & 13.58  & -8.92  & -6.89  \\
    CapRL-3B~\cite{caprl} & 1.33  & 30.67  & 20.67  & 17.56  \\
    InternVL3.5-38b~\cite{internvl3_5} & 0.83  & 35.58  & 21.17  & 19.19  \\
    \midrule
    Qwen3-VL-2B~\cite{qwen3_vl} & \underline{32.50}  &\underline{ 47.17}  & \underline{44.50 } & \underline{41.39}  \\
    CCCaption-2B (Ours) & \textbf{41.67 } & \textbf{49.17 } & \textbf{48.33 } & \textbf{46.39 } \\
    \bottomrule
\end{tabularx}%
    }
    \caption{Results of the CapArena evaluation~\cite{cheng2025caparena} across different captioning models. Bold numbers indicate the best performance, underlined numbers indicate the runner-up. Qwen3-VL-32B serves as the judge model. ``Claude'', ``CPMV'', and ``Cog2'' denote Claude-3.5-Sonnet-2024, MiniCPM-V2.6-8B, and CogVLM2-llama3-chat-19B, respectively.}
  \label{tab:caparena}%
\vspace{-12pt}
\end{table}

\noindent\textbf{Dynamic Query Sampling}. During training, we find that not all queries contribute equally to the gradient updates. In particular, some simple queries are answered correctly by almost all rollout captions. Under the GRPO framework, such queries yield tiny advantages and negligible gradients, which ultimately hurt training efficiency. Inspired by the dynamic sample selection strategy in DAPO~\cite{yu2025dapo}, we dynamically sample queries using the following formula:
\begin{equation}
    \tilde{\mathcal{Q}}
    = \bigl\{\,\mathbf{q} \in \mathcal{Q}_{\text{c}}
        \,\big|\,
        u_{\mathbf{q}} = 1,\;
        u_{\mathbf{q}} \sim \mathrm{Bernoulli}(c(\mathbf{q}))
      \bigr\},
\end{equation}
where $\mathcal{Q}_{\text{c}}$ is a set of candidate queries, $c(\mathbf{q}) \in [0,1]$ denotes the normalized contribution of query $\mathbf{q}$. The initial contribution is set as $1 - \frac{1}{m}\sum_{i=1}^m\mathbb{I}(M_{J}(\mathbf{x}, \mathbf{q}))$, where $m$ is the number of initial sampling. It is updated at each epoch by adding the variance of the rollout accuracy.

At the beginning of training, the contribution $c(\mathbf{q})$ is nearly the same for most queries, so the sampler samples them evenly. As training progresses, these uninformative queries (i.e., those that become almost always correct or incorrect) receive lower $c(\mathbf{q})$ values and are sampled less often, thereby increasing intra-group advantages and improving training efficiency.

\begin{table*}[t!]
  \centering
    \renewcommand{\arraystretch}{1.2}
    \scalebox{0.83}{
\begin{tabularx}{1.2\textwidth}{l|>{\centering\arraybackslash}X>{\centering\arraybackslash}X>{\centering\arraybackslash}X>{\centering\arraybackslash}X>{\centering\arraybackslash}X>{\centering\arraybackslash}X>{\centering\arraybackslash}X>{\centering\arraybackslash}X>{\centering\arraybackslash}X>{\centering\arraybackslash}X>{\centering\arraybackslash}X>{\centering\arraybackslash}X>{\centering\arraybackslash}X>{\centering\arraybackslash}X}
    \toprule
    \multicolumn{1}{c|}{\multirow{2}{*}{{MLLM}}} & ChartQA & CharXiv & InfoVQA & Verse & MMB   & MMMUPro & OCR   & COCO  & WM2Pro & \multicolumn{1}{c}{\multirow{2}{*}{{Avg.}}} \\
     & \cite{chartqa} & \cite{charxiv} & \cite{infovqa} & \cite{mathverse} & \cite{mmbench}   & \cite{mmmupro} & \cite{ocrbench}   & \cite{refcoco}  & \cite{WeMath20Pro} &  \\
    \midrule
    LLaVa-V1.6-34B~\cite{llava_v16} & 52.60  & 26.33  & 34.46  & 27.45  & 71.61  & 33.33  & 34.95  & 37.08  & 32.67  & 38.94  \\
    Qwen2.5-VL-3B~\cite{qwen2_5_vl} & 69.72  & 31.66  & 51.81  & 29.53  & 72.58  & 36.97  & 44.01  & 38.80  & 36.88  & 45.77  \\
    CapRL-3B~\cite{caprl} & 73.92  & 35.11  & \underline{55.94  }& 33.16  & 79.16  & 41.01  & 50.16  & 52.53  & 38.64  & 51.07  \\
    InternVL3.5-38b~\cite{internvl3_5} & 74.84  & \textbf{37.30 } & \textbf{59.40 } & 34.91  & 77.04  & 41.62  & \underline{50.81}  & 47.28  & 38.82  & 51.34  \\
    Qwen3-VL-32B~\cite{qwen3_vl} & \textbf{75.72 } & 35.74  & 51.94  & \textbf{35.82 } & \textbf{81.34 } & 43.64  & 49.19  & \textbf{56.82 } & \textbf{40.14 } & \underline{52.26}  \\
    \midrule
    Qwen3-VL-2B~\cite{qwen3_vl} & 71.00  & 31.66  & 51.78  & 33.23  &\underline{ 79.44 } & \underline{44.24 } & 50.16  & 50.41  & 38.58  & 50.05  \\
    CCCaption-2B (Ours) & \underline{75.12}  & \underline{36.99 } & 55.18  & \underline{35.11}  & 79.28  & \textbf{44.65 } & \textbf{55.34 } & \underline{53.70}  &\underline{ 39.83}  & \textbf{52.80 } \\
    \bottomrule

\end{tabularx}%
    }
  \caption{Results of the Prism evaluation~\cite{qiao2024prism} across different captioning models. Bold numbers indicate the best performance, underlined numbers indicate the runner-up.
  }
  \label{tab:vqa}%
  \vspace{-12pt}
\end{table*}

\begin{table*}[t!]
  \centering
    \renewcommand{\arraystretch}{1.2}
    \scalebox{0.83}{
\begin{tabularx}{1.2\textwidth}{l|>{\centering\arraybackslash}X>{\centering\arraybackslash}X>{\centering\arraybackslash}X>{\centering\arraybackslash}X>{\centering\arraybackslash}X>{\centering\arraybackslash}X>{\centering\arraybackslash}X>{\centering\arraybackslash}X>{\centering\arraybackslash}X>{\centering\arraybackslash}X>{\centering\arraybackslash}X>{\centering\arraybackslash}X>{\centering\arraybackslash}X>{\centering\arraybackslash}X}
    \toprule
    \multicolumn{1}{c|}{\multirow{2}{*}{{MLLM}}}   & AI2D  & ChartQA & Hallusion & MMB   & MME   & MMMU  & MMStar & OCR   & WM2Pro & \multicolumn{1}{c}{\multirow{2}{*}{{Avg.}}}  \\
      & \cite{ai2d}  & \cite{chartqa} & \cite{hallusionbench} & \cite{mmbench}   & \cite{mme}   & \cite{mmmu}  & \cite{mmstar} & \cite{ocrbench}   & \cite{WeMath20Pro} &  \\
    \midrule
    LLaVa-V1.6-34B~\cite{llava_v16} & 51.03  & 60.55  & 74.52  & 70.22  & 68.95  & 74.27  & 64.01  & 77.19  & 64.94  & 67.30  \\
    Qwen2.5-VL-3B~\cite{qwen2_5_vl} & 48.71  & 61.73  & 74.76  & 68.96  & 65.81  & 75.77  & 63.46  & 78.09  & 72.66  & 67.77  \\
    CapRL-3B~\cite{caprl} & 48.62  & 63.86  & 73.03  & 67.41  & 62.65  & 76.01  & 61.98  & 76.19  & 72.44  & 66.91  \\
    InternVL3.5-38b~\cite{internvl3_5} & 50.86  & 64.11  & 75.93  & \textbf{71.10 } & \textbf{68.15 } & 78.42  &\underline{ 65.32 } & \underline{82.72 } & \textbf{75.12 } & \underline{70.19}  \\
    Qwen3-VL-32B~\cite{qwen3_vl} & \textbf{52.38 } & \textbf{66.82 } & 75.45  & 69.20  & 64.38  & \underline{79.03}  & 64.37  & 82.35  & 74.87  & 69.87  \\
    \midrule
    Qwen3-VL-2B~\cite{qwen3_vl} & 51.51  & 64.53  & \textbf{77.51 } &\underline{ 70.91 } & \underline{66.78 } & 78.85  & 64.83  & 82.32  & 74.45  & \underline{70.19}  \\
    CCCaption-2B (Ours) & \underline{51.88}  &\underline{ 66.72}  & \underline{76.72}  & \underline{70.91}  & 66.16  & \textbf{79.07 } & \textbf{65.50 } & \textbf{83.09 } & \underline{75.10 } & \textbf{70.57 } \\
    \bottomrule

\end{tabularx}%
    }
  \caption{Results of the Hallucinations evaluation~\cite{lee2024toward} across different captioning models. Bold numbers indicate the best performance, underlined numbers indicate the runner-up.
  }
  \label{tab:hall}%
  \vspace{-16pt}
\end{table*}

\noindent\textbf{Overall}. 
Finally, we combine the two symmetric rewards into a single reward used for reinforcement learning:
\begin{equation}
    \mathcal{R}(\mathbf{x},\hat{\mathbf{y}})
    = \alpha\, \mathcal{R}_{\text{comp}}(\mathbf{x},\hat{\mathbf{y}})
      + (1-\alpha)\,\mathcal{R}_{\text{corr}}(\mathbf{x},\hat{\mathbf{y}}),
\end{equation}
where $\alpha \!\in\! [0,1]$ controls the trade-off. Substituting $\mathcal{R}(\mathbf{x},\hat{\mathbf{y}})$ into the RL objective in Eq.~\eqref{RL}, we obtain our CCCaption training objective for optimizing $M_\theta$.

\section{Experiments}

\begin{figure*}
    \centering
    \includegraphics[width=\linewidth]{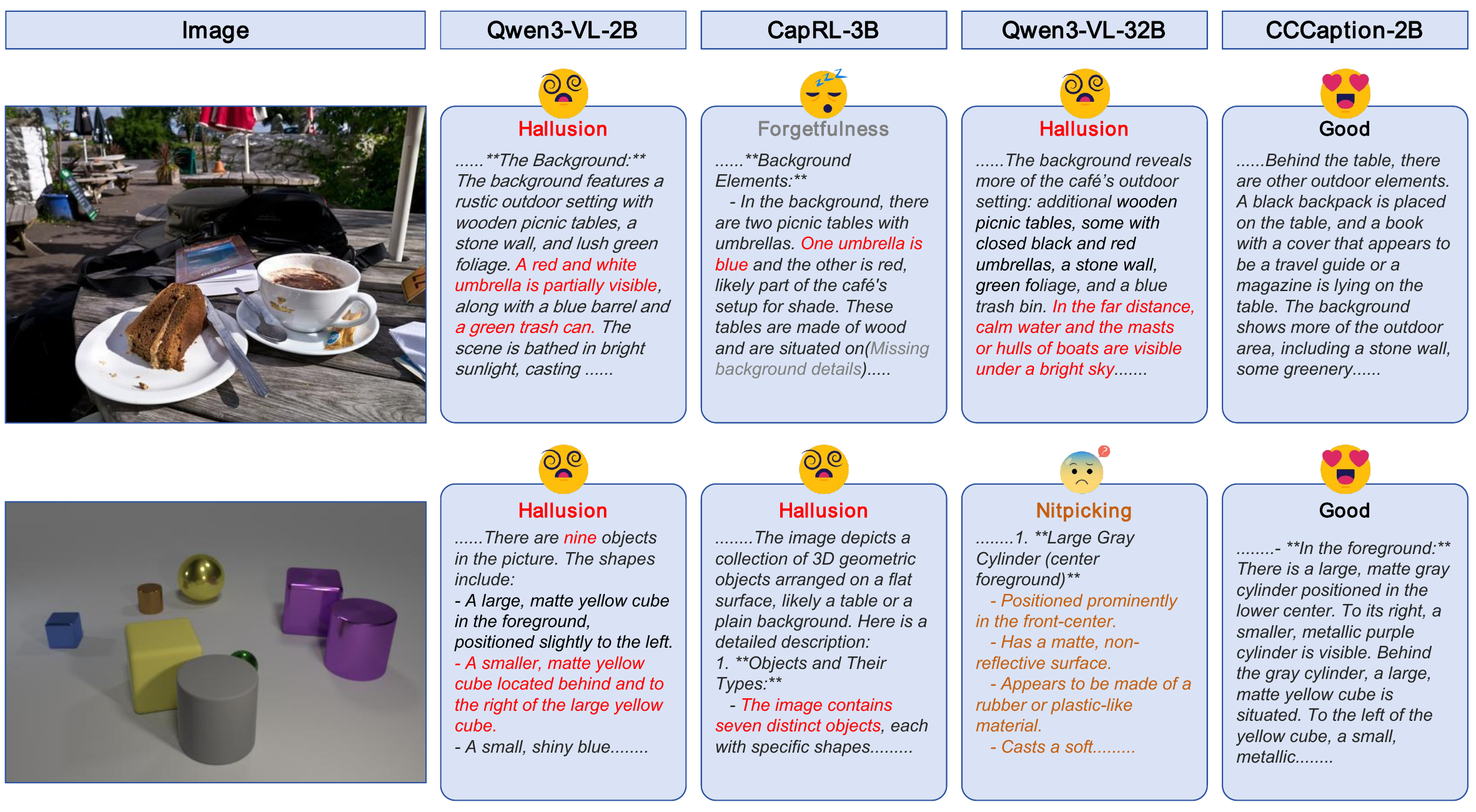}
    \caption{Case analysis across different captioning models. From left to right, the captioning models include Qwen3-VL-2B~\cite{qwen3}, CapRL-3B~\cite{caprl}, Qwen3-VL-32B~\cite{qwen3}, and CCCaption-2B (Ours), with image caption outputs for different queries labeled for hallucination, forgetfulness, and nitpicking. Both cases are derived from the MMBench dataset~\cite{mmbench}.}
    \label{fig:case}
    \vspace{-12pt}
\end{figure*}

\subsection{Evaluation Framework}
This work uses three frameworks to evaluate caption quality: the Prism, the Hallucinations, and the CapArena framework. Below, we will provide a detailed introduction to these frameworks and the associated datasets.

\noindent\textbf{Prism Framework}~\cite{qiao2024prism,caprl} evaluates caption quality based on the accuracy with which a judge model answers questions derived from the generated caption. Captions that answer more visual questions indicate that they cover more details of the images. So nearly all multimodal datasets can be used for caption quality evaluation in the Prism Framework. The primary datasets we use are as follows:
\begin{itemize}
    \item ChartQA (a dataset focused on evaluating chart comprehension, logical, and linguistic understanding)~\cite{chartqa}
    \item CharXiv (real-world charts extracted from research, designed to test MLLMs to interpret complex charts)~\cite{charxiv}
    \item InfoVQA (a Q\&A dataset featuring visual and textual elements, designed to test multimodal understanding)~\cite{infovqa}
    \item MathVerse (Verse, a benchmark for visual mathematical reasoning, covering multiple subjects, and diagrams)~\cite{mathverse}
    \item MMBench (MMB, a comprehensive multimodal understanding benchmark that spans multiple task types)~\cite{mmbench}
    \item MMMUPro (a benchmark for interdisciplinary understanding, covering a wide range of subject areas)~\cite{mmmupro}
    \item OCRBench (OCR, a benchmark dedicated to OCR, document-based Q\&A, and related tasks)~\cite{ocrbench}
    \item RefCOCO (COCO, a dataset designed to evaluate the object MLLM's localization ability in varied contexts)~\cite{refcoco}
    \item WeMath2.0-Pro (WM2Pro, a dataset specifically crafted for visual mathematical reasoning)~\cite{WeMath20Pro}
\end{itemize}

\noindent\textbf{Hallucinations Framework}~\cite{lee2024toward,feng2024more} splits the image caption into atomic captions, then uses MLLMs to detect whether these atomic captions contain hallucination. 
Additionally, we supplement the following datasets to detect hallucinations in captioning models:
\begin{itemize}
    \item AI2D (a dataset of diagrams from grade‑school science textbooks, richly annotated with linked questions)~\cite{ai2d}
    \item HallusionBench (an advanced diagnostic benchmark for VLMs that emphasises hallucination)~\cite{hallusionbench}
    \item MME (a comprehensive benchmark covering 14 subtasks of perception across multimodal scenarios)~\cite{mme}
    \item MMMU (a multimodal understanding benchmark that spans college‑level tasks across multiple subjects)~\cite{mmmu}
    \item MMStar (an elite vision‑indispensable benchmark comprising six core capabilities)~\cite{mmstar}
\end{itemize}

\noindent\textbf{CapArena}~\cite{cheng2025caparena} constructs an ``arena-style'' evaluation platform, consisting of 600 pairwise caption battles, with each struggle judged by a MLLM, which, using in this study, is Qwen3-VL 32B. In these battles, the candidate model’s caption is compared to that of a baseline model, and the judge assigns a score of +1, 0, or -1 depending on whether the candidate caption is judged as better, the same, or worse, respectively. The baseline models used in this study include Claude‑3.5‑Sonnet‑2024~\cite{anthropic2024claude}, MiniCPM‑V2.6‑8B~\cite{huggingface2024minicpm}, and CogVLM2‑llama3‑chat‑19B~\cite{huggingface2024cogvlm2}. The candidate model’s overall score is the normalised sum of scores across all pairwise comparisons. This scoring method enables detailed caption comparison, and the resulting rankings are highly correlated with human preferences~\cite{cheng2025caparena}.

\subsection{Implementation Details}
\noindent\textbf{CCaption-44k}. We collected 140k images from 16 public datasets. After deduplication and removal of anomalous data (e.g., resolution issues, image contamination, etc.), we obtained a final dataset of 44k samples. The MLLM list used for query sampling includes LLaVa-V1.6 34B~\cite{llava_v16}, Qwen2.5-VL 72B~\cite{qwen2_5_vl}, InternVL3.5 38B~\cite{internvl3_5}, and Qwen3-VL 32B~\cite{qwen3_vl}. The query sampling algorithm is shown in Algorithm~\ref{alg:qa_pipline} with threshold $\tau=0.1$. In practice, we sample 15 non-overlapping queries per LVLM at once and use the pipeline in Figure~\ref{fig:method}(b) to filter low-quality queries. The embedding model for both images and queries is Ops-MM-embedding~\cite{ops_mm_embedding_v1_7B}. The final CCaption-44k training dataset covers images comprehensively, with an average of 10 queries per sample.

\noindent\textbf{Training Setting}. The base model we trained is Qwen3-VL 2B~\cite{qwen3_vl}, and the judge model for completeness reward is Qwen2.5-VL 3B~\cite{qwen2_5_vl}. The dynamic query sampling strategy is used only for completeness reward, sampling 5 queries per step. The judge model for correctness reward is Qwen3-VL 2B, with a maximum of 5 sub-caption queries allowed per caption. Due to the sparsity of hallucinations, the weight for hybrid rewards is set low at $\alpha=0.05$. 
We use the reinforcement learning training framework of GRPO~\cite{guo2025deepseek}, implemented with VeRL~\cite{sheng2024hybridflow}. The rollout is set to 5, the learning rate is 1e-6, and the number of epochs is 20.
Both training and judge model deployments were conducted on NVIDIA H100 GPUs.

\noindent\textbf{Evaluation Setting}. For captioning models, we evaluated state-of-the-art LVLMs, including LLaVa-V1.6 34B, Qwen2.5-VL 3B and 72B, CapRL 3B~\cite{caprl}, InternVL3.5 38B, Qwen3-VL 2B and 32B, as well as our CCCaption-2B. The evaluation frameworks and datasets have been introduced above. All evaluation experiments were conducted on NVIDIA H100 GPUs. Additional implementation details are provided in the supplementary materials.

\subsection{Result Analysis}
\noindent\textbf{Q1.} \textbf{\textit{How does CCCaption-2B perform?} }

\noindent\textbf{A1.} 
CCCaption achieves state-of-the-art (SOTA) performance across multiple evaluation frameworks. For the CapArena evaluation, as shown in Table~\ref{tab:caparena}, CCCaption-2B significantly outperforms Claude-3.5-Sonnet-2024 in captioning. Compared to the base model Qwen3-VL 2B, it shows an improvement of 5.00 (12.08\%), and compared to CapRL 3B, it improves by 28.83 (164.24\%).
For Prism evaluation, as shown in Table~\ref{tab:vqa}, CCCaption-2B improves by 
1.72 (3.38\%) compared to CapRL 3B. Despite being a smaller model, CCCaption-2B performs comparably to Qwen3-VL 32B (i.e., CCCaption-2B outperforms it by 0.54). Notably, the model shows a significant improvement on OCRBench, achieving 55.34, surpassing the second-place InternVL3.5 38B by 4.53 (8.91\%).

For Hallucinations evaluation, as shown in Table~\ref{tab:hall}, CCCaption-2B improves by 0.38 (0.54\%) compared to the base model Qwen3-VL 2B. In contrast, CapRL, lacking the correctness reward mechanism, shows a significant increase in hallucination compared to the base model Qwen2.5-VL 3B, with the correctness metric decreasing by 0.86 (1.27\%). At the same time, CCCaption-2B demonstrates higher correctness than Qwen3-VL 32B, improving by 0.70 (1.00\%).

\begin{figure}
    \centering
    \includegraphics[width=\linewidth]{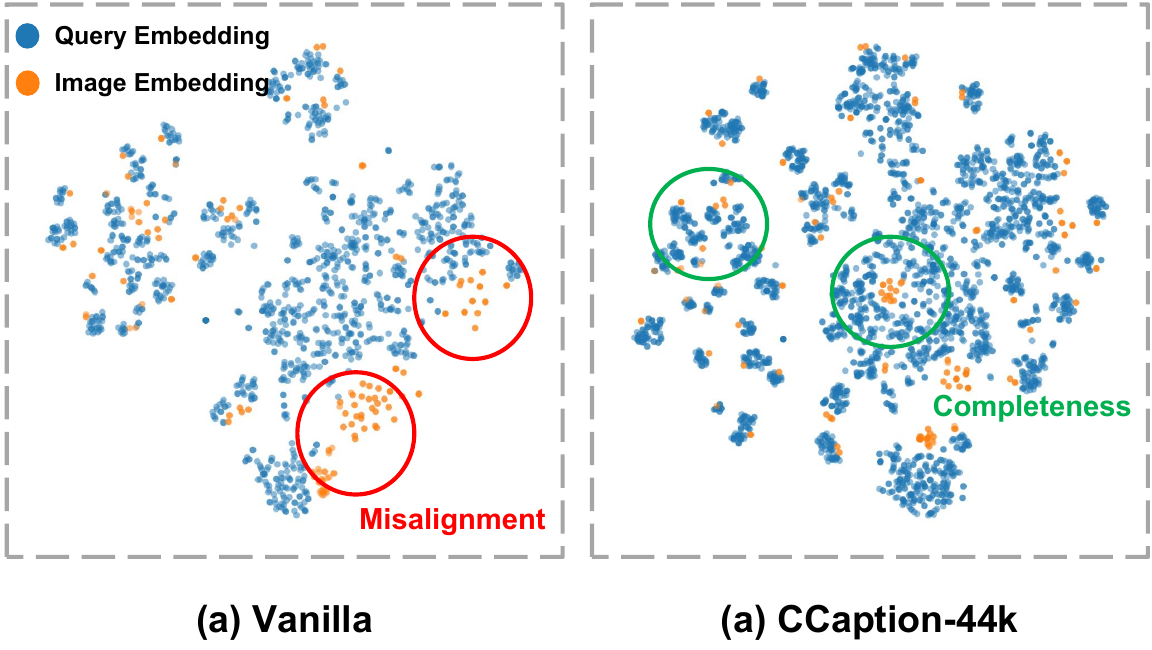}
    \caption{Query and image embedding scatters across different datasets. ``Vanilla'' refers to queries generated by a single MLLM, using the CapRL method~\cite{caprl}, while ``CCaption-44k'' represents our approach, which utilizes multiple MLLMs for generation and employs a diversity metric to measure query diversity. The embedding model used is Ops-MM-embedding~\cite{ops_mm_embedding_v1_7B}, with dimensionality reduction performed using t-SNE~\cite{maaten2008visualizing}.}
    \label{fig:image_qa_distribution}
    \vspace{-16pt}
\end{figure}

\noindent\textbf{Q2.} \textbf{\textit{Are the queries in CCaption-44k complete and effective enough?} }

\noindent\textbf{A2.} 
Queries in CCaption-44k are complete. As shown in Figure~\ref{fig:image_qa_distribution}, we visualized the embeddings of images and queries using t-SNE~\cite{maaten2008visualizing}. The figure illustrates that ``Vanilla'' (i.e., single MLLM-generated queries according to CapRL~\cite{caprl}) does not adequately capture image features. There are regions where query embeddings misalign with image features, while these regions are crucial for captioning. In contrast, CCaption-44k (i.e., multi-MLLM-generated queries with high diversity) covers almost all image features, yielding a more complete query set.
 
Queries in CCaption-44k are effective. As shown in Table~\ref{tab:ablation_data}, the model trained on the CCaption-44k outperforms the base model by 2.69 (6.2\%) and the model trained on ``Vanilla'' queries by 1.87 (4.25\%). Furthermore, CCaption exhibits a strong scaling law, with model performance improving as the training dataset grows. Specifically, with 44k samples, the model’s performance improved by 1.64 and 1.01 compared to 8k and 16k samples, respectively.

\noindent\textbf{Q3.} \textbf{\textit{How does the dynamic query sampling strategy improve model training efficiency?} }

\begin{figure}
    \centering
    \includegraphics[width=\linewidth]{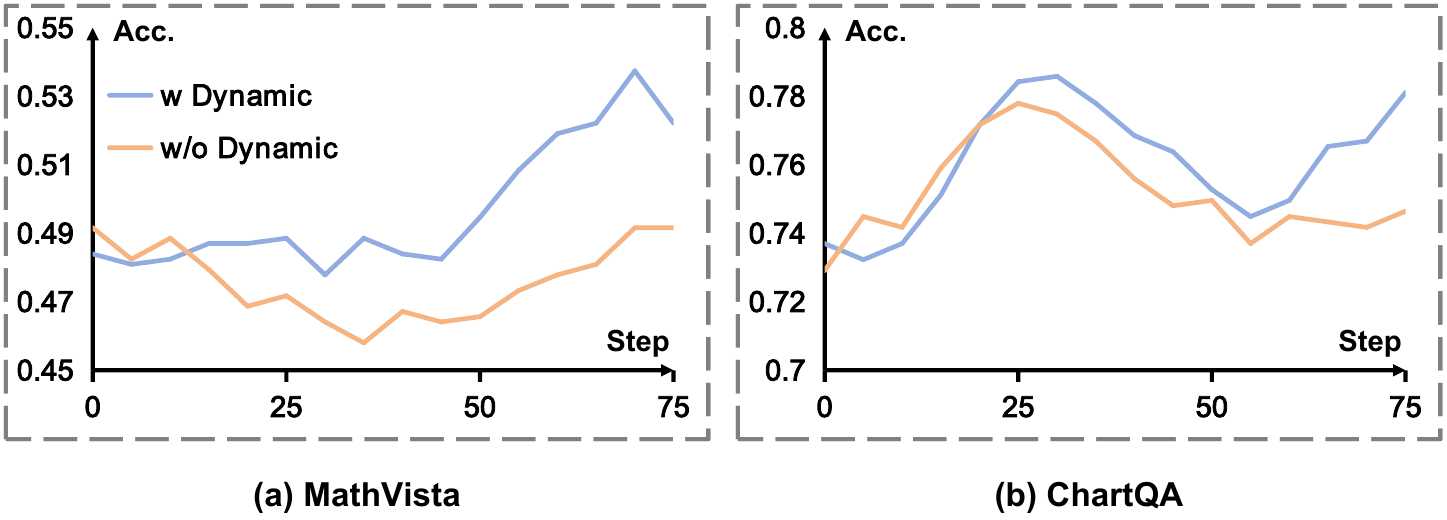}
    \caption{Performance whether the dynamic query sampling strategy is used during training. ``w Dynamic'' denotes the use of the strategy, while ``w/o Dynamic'' indicates the absence of the strategy. ``Acc.'' refers to the accuracy under the Prism evaluation~\cite{qiao2024prism}.}
    \label{fig:dynamic_qa}
    \vspace{-12pt}
\end{figure}

\noindent\textbf{A3.} 
The dynamic query sampling strategy significantly improves training efficiency. As shown in Figure~\ref{fig:dynamic_qa}, whether the strategy is implemented during the early stages of training has little impact on performance for datasets such as MathVista or ChartQA. However, as training progresses, models using the strategy show significant performance improvements compared to those without it. This indicates that the dynamic query sampling strategy facilitates model convergence and effectively reduces training time.

Next, we discuss how the dynamic query sampling strategy enhances training efficiency. In fact, the difficulty of obtaining an atomic semantic fact for an image varies, and the difficulty of visual queries depends on the facts. For example, simple queries can help the model determine the gradient convergence direction early on. However, as training progresses, all rollout captions correctly answer these simple queries at each step, significantly reducing the group advantage and leading to a notable decrease in the policy model's gradient and, consequently, training efficiency. Similarly, more difficult queries can also reduce training efficiency. The introduced dynamic query sampling strategy dynamically adjusts the sampling probability for each query in each sample based on its contribution to the reward. This strategy ensures that ``controversial'' queries are used to calculate the reward in the later stages of training, preventing a reduction in group advantages and stabilizing the policy model's gradient.

\begin{table}[t!]
  \centering
    \renewcommand{\arraystretch}{1.2}
    \scalebox{0.9}{
\begin{tabularx}{0.53\textwidth}{l|>{\centering\arraybackslash}X>{\centering\arraybackslash}X>{\centering\arraybackslash}X>{\centering\arraybackslash}X>{\centering\arraybackslash}X>{\centering\arraybackslash}X}
    \toprule
    \multicolumn{1}{c|}{\multirow{2}{*}{{Training Data}}}
    & CharXiv & InfoVQA & MMMU  & MMStar & \multicolumn{1}{c}{\multirow{2}{*}{{ Avg.}}} \\
    & \cite{charxiv} & \cite{infovqa} & \cite{mmmu}  & \cite{mmstar} & \\
    \midrule
    w/o RL  & 31.66  & 51.78  & 39.64  & 49.93  & 43.25  \\
    Vanilla & 32.92  & 51.94  & \underline{40.96}  & 50.47  & 44.07  \\
    CCaption-8k  & 35.42  & \underline{53.23 } & 40.80  & 47.73  & 44.30  \\
    CCaption-16k & \underline{35.74}  & 52.91  & 40.29  & \textbf{50.80 } & \underline{44.94 } \\
    CCaption-44k & \textbf{36.99 } & \textbf{55.18 } & \textbf{41.07 } & \underline{50.53}  & \textbf{45.94 } \\
    \bottomrule
\end{tabularx}%
    }
    \caption{Ablation of training data under the Prism evaluation~\cite{qiao2024prism}. Bold numbers indicate the best performance, and underlined numbers indicate the runner-up. ``Vanilla'' refers to the queries generated according to the CapRL method~\cite{caprl}, while ``CCaption'' represents our approach. Qwen3-VL 2B~\cite{qwen3_vl} is used for training.
}
  \label{tab:ablation_data}%
\vspace{-12pt}
\end{table}

\noindent\textbf{Q4.} \textbf{\textit{Does the correctness suppress hallucinations?} }

\noindent\textbf{A4.} 
From the quantitative results in Table~\ref{tab:hall}, CapRL 3B without the introduction of correctness rewards exhibits a significant increase in hallucination compared to its base model, Qwen2.5-VL 3B, with a performance decrease of 0.86 (1.27\%). In contrast, CCCaption-2B shows an improvement of 0.38 (0.54\%) over its base model, Qwen3-VL 2B, further suppressing hallucinations in the base model. 
From the qualitative results in Figure~\ref{fig:case}, CCCaption-2B provides more accurate captions. For example, in the second row, both Qwen3-VL 2B and CapRL 3B incorrectly report the number of objects. Qwen3-VL 2B hallucination described ``the yellow cube behind'', while CapRL 3B describes only seven objects. In contrast, CCCaption-2B accurately identifies eight objects and provides a correct description of their relationships.

Correctness rewards are achieved by splitting long captions into sub-caption queries that contain atomic facts. The judge model excels at evaluating these short queries, providing high-confidence scores. These accurate hallucination scores serve as rewards that effectively encourage the model to avoid generating incorrect captions, thereby significantly suppressing hallucinations.

\noindent\textbf{Q5.} \textbf{\textit{Are all components of the CCCaption necessary?}}

\noindent\textbf{A5.} 
The components in the CCCaption framework are effective. \textbf{Q3.} and \textbf{Q4.} analyze how the dynamic query sampling strategy significantly improves training efficiency, and how the correctness reward effectively mitigates hallucinations. We also conducted extensive ablation experiments, as shown in Table~\ref{tab:ablation_rl}. Our performance improved by 0.91 (1.32\%) compared to ``w/o Dynamic'', demonstrating that the dynamic query sampling strategy enhances training efficiency. Using only the completeness~(``w/o Corr'') or correctness~(``w/o Comp'') reward leads to a decrease in performance on the other metric (e.g., under OCR, using only completeness resulted in a 0.39 performance drop compared to the base model in the Hallucinations evaluation, while using only correctness caused a 9.71 drop in performance compared to the base model in the Prism evaluation), indicating that captions cannot be both complete and correct when using only one reward. The hybrid reward improved performance by 2.51 (3.7\%) and 4.13 (6.27\%) over the completeness and correctness-only rewards, respectively.

\begin{table}[t!]
  \centering
    \renewcommand{\arraystretch}{1.2}
    \scalebox{0.9}{
\begin{tabularx}{0.53\textwidth}{l|>{\centering\arraybackslash}X>{\centering\arraybackslash}X>{\centering\arraybackslash}X>{\centering\arraybackslash}X>{\centering\arraybackslash}X>{\centering\arraybackslash}X}
    \toprule
    \multicolumn{1}{c|}{\multirow{2}[4]{*}{Component}} & \multicolumn{2}{c}{ChartQA~\cite{chartqa}} & \multicolumn{2}{c}{OCR~\cite{ocrbench}} & \multirow{2}[4]{*}{Avg.} \\
\cmidrule{2-5}          & Prism & Hall. & Prism & Hall. &  \\
    \midrule
    w/o RL   & 71.00  & 64.53  & 50.16  & 82.32  & 67.00  \\
    w/o Corr & \underline{74.56}  & 65.84  & 47.90  & 81.93  & 67.56  \\
    w/o Comp & 71.60  & \textbf{68.11 } & 40.45  & \underline{83.55}  & 65.93  \\
    w/o Dynamic & 73.92  & 66.33  & \underline{52.43}  & \textbf{83.92 } & \underline{69.15 } \\
    Ours &     \textbf{75.12 } & \underline{66.72}  & \textbf{55.34 } & 83.09  & \textbf{70.07 } \\
    \bottomrule
\end{tabularx}%
    }

\caption{Ablation of key components in the CCCaption framework. 
``Prism'' and ``Hall.'' refer to the results under the Prism~\cite{qiao2024prism} and Hallucinations~\cite{lee2024toward} evaluations, respectively. ``w/o Corr'' and ``w/o Comp'' denote the use of only completeness and correctness rewards, respectively. ``w/o Dynamic'' indicates without the dynamic sampling strategy. Qwen3-VL 2B~\cite{qwen3_vl} is used for training.}
  \label{tab:ablation_rl}%
\vspace{-12pt}
\end{table}

\section{Conclusion}

In this paper, we propose the CCCaption framework, which addresses key challenges in image captioning by focusing on two essential properties: completeness and correctness. To achieve this, we integrate a dual-reward mechanism that optimizes both visual information coverage and caption accuracy. 
Another crucial component of our framework is the dynamic query sampling strategy, which significantly enhances training efficiency by adjusting the probability of sampling diverse queries. 
Furthermore, we use a multi-MLLMs query generation pipeline to construct the CCaption-44k dataset, comprising 44,077 samples that comprehensively cover key visual information in images. 
Building upon these advancements, we trained the CCCaption-2B model based on Qwen3-VL 2B. Extensive experiments demonstrate that CCCaption-2B outperforms existing models, including the more powerful Qwen3-VL 32B. CCCaption-2B achieves a new state-of-the-art in image captioning. 
As for future work, we aim to refine our evaluation metrics for correctness and completeness, potentially introducing more nuanced indicators that better capture the subtleties of these properties.

\section*{Acknowledgments}
This work was supported by the Strategic Priority Research Program of the Chinese Academy of Sciences under Grant No. XDA0460205. Thanks to Shopee for providing computing power support.

{
    \small
    \bibliographystyle{ieeenat_fullname}
    \bibliography{main}
}


\end{document}